\documentclass{article} 
\usepackage{iclr2027_conference,times}

\usepackage{amsmath,amsfonts,bm}

\def\Figref#1{Figure~\ref{#1}}

\def\eqref#1{equation~\ref{#1}}

\def\Algref#1{Algorithm~\ref{#1}}

\def\1{\bm{1}}

\DeclareMathAlphabet{\mathsfit}{\encodingdefault}{\sfdefault}{m}{sl}
\SetMathAlphabet{\mathsfit}{bold}{\encodingdefault}{\sfdefault}{bx}{n}

\usepackage{hyperref}
\hypersetup{hypertexnames=false}
\usepackage{url}
\usepackage{graphicx}
\usepackage{algorithm}
\usepackage{algpseudocode}
\usepackage{booktabs}
\usepackage[table]{xcolor}
\usepackage{subcaption}

\newcommand{\na}{\textemdash}

\title{Memory Forcing: Attendable Mid-Horizon History for Streaming Video Generation}

\author{
Jiaming Zhang$^{1,*,\dagger}$,
Xinyu Wang$^{2,*,\dagger}$,
Huafeng Shi$^{3}$,
Gangshan Wu$^{1}$,
Limin Wang$^{1,4,\ddagger}$ \\
$^{1}$State Key Laboratory for Novel Software Technology, Nanjing University \\
$^{2}$Tsinghua University \\
$^{3}$Kling Team, Kuaishou Technology \\
$^{4}$Shanghai Artificial Intelligence Laboratory
}

\iclrfinalcopy 
\begin{document}

\maketitle
\lhead{arxiv preprint}
\begingroup
\renewcommand{\thefootnote}{\fnsymbol{footnote}}
\footnotetext[1]{Equal contribution.}
\footnotetext[2]{This work was conducted during the authors' internships at Kling Team, Kuaishou Technology.}
\footnotetext[3]{Corresponding author: \texttt{lmwang@nju.edu.cn}.}
\endgroup

\begin{abstract}
Autoregressive video diffusion enables causal video streaming without a bidirectional pass over the full clip, but existing few-step systems usually retain only the opening and most recent frames in a fixed-size KV cache.
Once an event leaves this window, later frames can no longer attend to it, a failure we term \emph{mid-horizon forgetting}.
We present \emph{Memory Forcing}, a few-step streaming method that preserves this missing history without increasing the cache size.
Its \emph{Archive \& Working Banks} partition the cache into sink, archive, and working regions, retaining diverse intermediate events alongside recent motion under fixed memory.
Because absolute temporal indices drift outside the training range, \emph{Bank-aware RoPE} reassigns indices at attention time so each bank remains distinguishable.
At 1.3B, Memory Forcing leads on longer clips, shows the smallest drop from 5s to 60s among methods reporting all four lengths, and preserves subjects and scenes through leave-and-return.
The same design scales to Wan2.2 5B, producing more physically plausible, realistic, and dynamic videos and, to our knowledge, the first public 5B model on this forcing line.
\end{abstract}

\section{Introduction}
\label{sec:intro}

Modern video diffusion models~\citep{ma2025controllable, zhang2026kvpo, wang2026subjectanchor, ma2026group, zhang2026zo3t, zhang2025mind, zhang2026robostereo, he2024co, chen2026human, he2025magicman, huang2026unityvideo, fang20263d} generate short clips with rich detail and coherent motion by denoising the full sequence with bidirectional attention.
These offline generators set a high quality bar but cannot emit a frame until the clip is finished.
Streaming applications instead need a causal generator with low latency and temporal consistency over long horizons.
CausVid~\citep{causvid2025} and Self-Forcing~\citep{selfforcing2025} distill a bidirectional teacher into a few-step causal student and can produce frames on the fly.
Their strictly causal prediction lets each frame inherit errors from its predecessors, so defects compound into drift.
Rolling Forcing~\citep{rollingforcing2026} mitigates that in-window drift with rolling-window joint denoising, so frames that share a window stay aligned.

Streaming video generation~\citep{wang2026liveedit} conditions each new chunk on frames already produced.
Those frames are stored as keys and values in a KV cache, which later denoising steps attend without re-encoding the past.
This cache keeps attention cost from growing with clip length.
Most current systems manage this cache as a recent FIFO of fixed length.
New chunks are appended, and the oldest keys are dropped when the window is full.
Local motion is preserved, but events that have left the window are gone.
A subject that leaves and later returns, or a scene that cuts away and comes back, is no longer addressable.
We call this \emph{mid-horizon forgetting}, where lost events sit between the opening of the clip and the recent FIFO window, so later queries have no slot to retrieve them.

To keep those mid-horizon events attendable, we present \emph{Memory Forcing}, a few-step streaming method that keeps rolling and DMD.
The opening of a clip, the recent motion, and the events in between are different kinds of history.
A single FIFO keeps only the recent keys, so intermediate events have nowhere to go.
We introduce \textbf{Archive \& Working Banks} and partition the KV cache into \emph{sink}, \emph{archive}, and \emph{working} banks.
New chunks enter working.
When working is full, Similarity writes a chunk into archive, and a full archive drops the oldest chunk by FIFO.
Mid-horizon events remain attendable, local motion stays in the recent window, and the cache does not grow.

Those banks must be placed on a time axis so later queries can tell the opening, the recent window, and mid-horizon events apart.
Previous methods write keys with \emph{Absolute} time $\tau^{\mathrm{abs}}$, the frame index at write time.
On a long video those indices grow outside the trained range, so later queries read stored events in the wrong place.
The same rule also fails a three-bank cache.
Archive is older than the recent window, and the three roles need different places on the axis, which one growing index cannot give them.
We introduce \textbf{Bank-aware RoPE}, store raw keys for the whole window, and re-assign a temporal index~$\tau$ before every attention.
The assigned indices stay in the trained range, the three banks remain distinguishable, and later queries read stored events in the right place.

We evaluate the same method at Wan2.1-T2V-1.3B and Wan2.2 5B~\citep{wan2025}.
1.3B matches prior few-step models and covers the main table and ablations that turn off the partition or $\tau$.
To our knowledge, the 5B checkpoint is the first public 5B on this forcing line and holds up better on crowded multi-subject scenes.
Experiments show Memory Forcing tops longer clips, drops least from 5s to 60s among methods reporting all four lengths, and still retrieves leave-and-return.

\begin{figure}[t]
  \centering
  \includegraphics[width=0.85\linewidth]{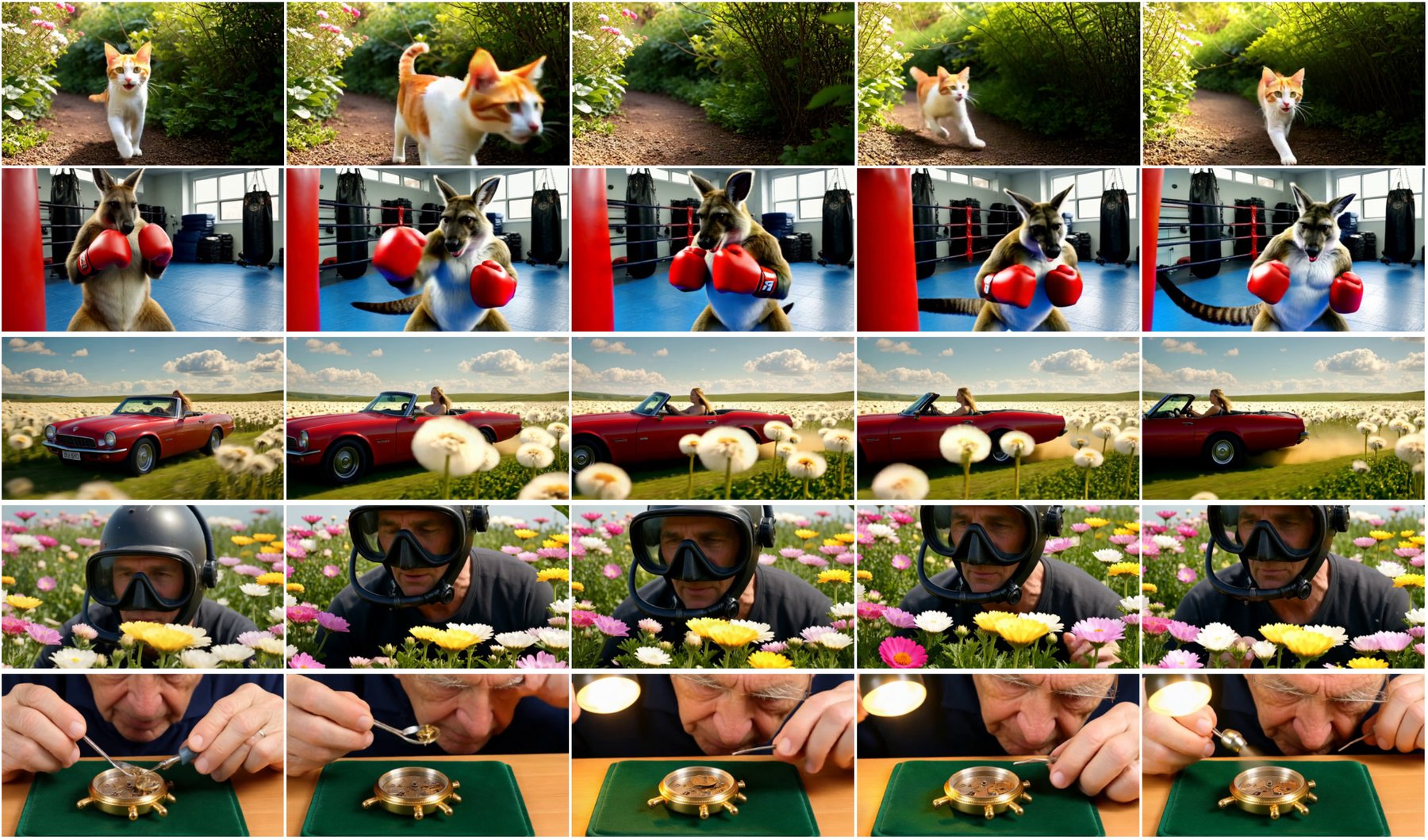}
  \vspace{-3mm}
  \caption{Qualitative samples from Memory Forcing.
  Each row is one generated clip, frames in time order.
  Subjects, motion, and scene layout stay consistent across the clip.}
  \label{fig:teaser}
  \vspace{-6mm}
\end{figure}

The contributions of this work are three.
First, we introduce Archive \& Working Banks, a sink / archive / working cache that keeps mid-horizon events attendable, with Similarity as the default write into archive.
Second, we introduce Bank-aware RoPE, which stores raw keys for the whole window and re-assigns a temporal index before every attention.
The three banks can then be told apart, and later queries read stored events in the right place.
Third, experiments show Memory Forcing tops longer clips, drops least from 5s to 60s among methods reporting all four lengths, and still retrieves leave-and-return.
To our knowledge, we also introduce the first public 5B.

\section{Related Work}
\label{sec:related}

\paragraph{Bidirectional video diffusion.}
Video diffusion models denoise the full clip with bidirectional attention.
Latent U-Nets such as SVD~\citep{svd2023} and VideoLDM~\citep{videoldm2023} set that recipe.
DiT-scale generators set the quality bar: Sora~\citep{sora2024}, MovieGen~\citep{moviegen2024}, CogVideoX~\citep{cogvideox2025}, HunyuanVideo~\citep{hunyuanvideo2025}, and Wan~\citep{wan2025}, but cannot emit a causal stream.
Autoregressive models such as NOVA~\citep{nova2025}, SANA-Video~\citep{sanavideo2025}, and InfinityStar~\citep{infinitystar2025} emit frames in time order and can grow a clip without a bidirectional pass over the full sequence.

\paragraph{Forcing, from causal distillation to a two-bank cache.}
CausVid~\citep{causvid2025} distills a bidirectional teacher with DMD~\citep{yin2024dmd}.
Self-Forcing~\citep{selfforcing2025} trains on self-generated history.
CausalForcing~\citep{causalforcing2026} uses a causal teacher ODE.
SkyReels-V2~\citep{skyreelsv22025} follows Diffusion Forcing~\citep{diffusionforcing2024}.
Rolling Forcing~\citep{rollingforcing2026} adds rolling-window repair, an attention sink, and a recent FIFO $L_{\mathrm{tem}}$.
LongLive~\citep{longlive2026} is scored in the same table.
Rolling Forcing holds in-window drift down, and the KV cache is still a sink plus $L_{\mathrm{tem}}$, so mid-horizon events fall out of the FIFO.

\paragraph{Long-range memory and position.}
RewardForcing~\citep{rewardforcing2026}, DeepForcing~\citep{deepforcing2026}, and RollingSink~\citep{rollingsink2026} strengthen only the sink, so intermediate history still leaves the window.
Infinity-RoPE~\citep{infinityrope2026} and MemRoPE~\citep{memrope2026} rewrite time or fold the past into fixed tokens, with no archive.
LongLive~\citep{longlive2026} and AnchorForcing~\citep{anchorforcing2026} recache on prompt changes, not leave-and-return.
DummyForcing~\citep{dummyforcing2026} and Forcing-KV~\citep{forcingkv2026} compress the window, not mid-horizon slots.
These still leave mid-horizon events without a slot, so we use Archive \& Working Banks to keep them attendable and Bank-aware RoPE so later queries read them in the right place.

\section{Method}
\label{sec:method}

Memory Forcing keeps the Rolling Forcing~\citep{rollingforcing2026} training recipe, its rolling schedule, and its DMD objective.
What changes is the KV cache later steps attend and the temporal index used once that cache is split by role.
Sections~\ref{sec:method:prelim}--\ref{sec:method:rolling} state the inherited generator, the DMD gradient, and rolling-window joint denoising.
Sections~\ref{sec:method:memory}--\ref{sec:method:scale} specify the sink, archive, and working banks, Bank-aware RoPE, Aligned cache insertion, and the same recipe on Wan2.2 5B.

\subsection{Preliminaries}
\label{sec:method:prelim}

\paragraph{Autoregressive video diffusion.}
Autoregressive video diffusion combines temporal causality with iterative denoising, so each new frame may look only at history that has already been produced.
For a clip $x_{1:N}=(x_1,\ldots,x_N)$ the joint therefore factorizes as
\begin{equation}
\label{eq:chain}
  p(x_{1:N})=\prod_{i=1}^{N} p(x_i\mid x_{<i}),
\end{equation}
and each factor is realized by denoising Gaussian noise conditioned on previously generated frames.
Streaming systems generate these frames in small chunks, reuse a temporal key--value cache of recent latents and a small sink of the opening frames, and condition on the current text embedding~\citep{causvid2025,anchorforcing2026,selfforcing2025,rewardforcing2026}.
One step may emit a chunk rather than a singleton, still called a frame unless size matters.

Those conditionals depend on how history is constructed.
Two choices are Teacher Forcing (TF) and Diffusion Forcing (DF)~\citep{diffusionforcing2024}.
Under TF the conditional for the $i$th frame at noise level $t_j$ is $p(x_i^{t_j}\mid x_{<i}^{0})$, so every history frame is clean.
Under DF it is $p(x_i^{t_j}\mid x_{<i}^{t\geq 0})$, so the histories are ground-truth frames corrupted with independent noise.
Training sees ground-truth prefixes while inference feeds the model's own predictions, known as exposure bias~\citep{schmidt2019}.
Self-Forcing~\citep{selfforcing2025} lets $G_{\theta}$ denoise each noisy frame given rolled-out $\hat{x}_{<i}^{0}$,
\begin{equation}
\label{eq:sf}
  p(\hat{x}_i^{t_j}\mid \hat{x}_{<i}^{0}),
  \qquad
  \hat{x}_i^{t_{j-1}}
  =
  \Psi\bigl(G_{\theta}(\hat{x}_i^{t_j}, t_j, \hat{x}_{<i}^{0}), t_{j-1}\bigr),
\end{equation}
with $\hat{x}_i^{t_T}\sim\mathcal{N}(0,I)$.
Here $\Psi$ injects Gaussian noise at the next lower noise level.
Training and inference then see the same exposure, so the student denoises on top of its own errors rather than a ground-truth prefix.
Relative to TF and DF, the train-time conditionals stay closer to a real rollout.

\paragraph{Distribution matching distillation.}
A self-rollout has no paired ground-truth frame, so a pixel denoising loss cannot be formed.
CausVid~\citep{causvid2025} distills a bidirectional multi-step teacher into a few-step causal student with Distribution Matching Distillation (DMD)~\citep{yin2024dmd}.
DMD minimizes a reverse KL between the student's output distribution $p_{\mathrm{gen},t}$ and a smoothed data distribution $p_{\mathrm{data},t}$ at sampled times $t$, as the difference of two scores.
\begin{equation}
\label{eq:dmd}
\begin{aligned}
  \nabla_{\theta}\mathcal{L}_{\mathrm{DMD}}
  &=
  \mathbb{E}_{t}\bigl[\nabla_{\theta}\mathrm{KL}(p_{\mathrm{gen},t}\,\Vert\,p_{\mathrm{data},t})\bigr] \\
  &\approx
  -\mathbb{E}_{t}
  \int
  \bigl(
    s_{\mathrm{data}}(\Psi(G_{\theta}(\epsilon),t),t)
    -
    s_{\mathrm{gen}}(\Psi(G_{\theta}(\epsilon),t),t)
  \bigr)
  \frac{\mathrm{d}G_{\theta}(\epsilon)}{\mathrm{d}\theta}\,
  \mathrm{d}\epsilon,
\end{aligned}
\end{equation}
where $\Psi$ is the forward process, $\epsilon$ is Gaussian noise, $G_{\theta}$ is the generator, and $s_{\mathrm{data}}$ and $s_{\mathrm{gen}}$ score the data and the generator's outputs.
The data score comes from a frozen teacher, the student score from a separate network.
Then \eqref{eq:sf} trains in few steps with this DMD and flow critic.

\begin{figure}[t]
  \centering
  \includegraphics[width=\linewidth]{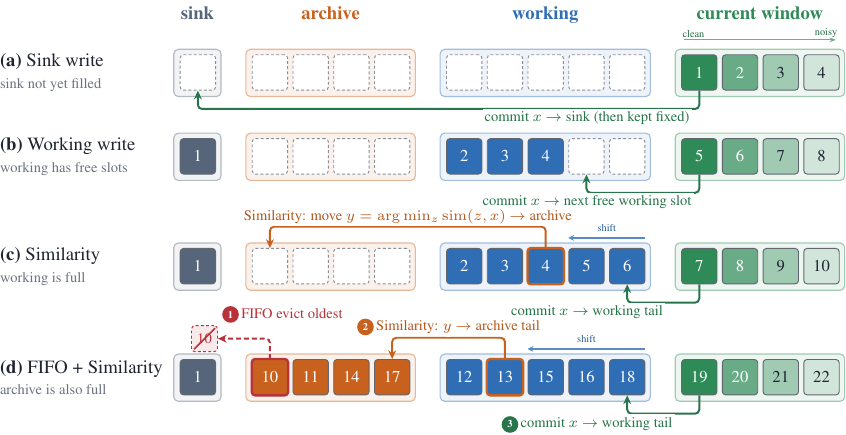}
  \vspace{-8mm}
  \caption{Archive \& Working Banks, same cache each row.
  (a) Only the current window is filled.
  (b) The first chunk goes into sink and stays.
  (c) Later chunks enter working while slots remain.
  (d) When working is full, Similarity writes into archive.
  (e) FIFO then drops the oldest archive chunk.}
  \label{fig:memory}
  \vspace{-5mm}
\end{figure}

\subsection{Rolling-window joint denoising}
\label{sec:method:rolling}

Self-Forcing with DMD aligns train and inference history through \eqref{eq:sf}.
Each step still denoises one frame against a clean prefix, and that frame never attends back into its history.
A local error is written into the next prefix and cannot be revised, so later frames inherit and compound it.

Rolling Forcing~\citep{rollingforcing2026} replaces that singleton with a window of $T$ frames that share one forward pass and rise in noise from old to new.
Neighbors revise a local error before commit, while each step emits one clean chunk.
Starting at frame $i$, one roll updates that window by
\begin{equation}
\label{eq:rf}
  p_{\theta}(\hat{x}_{i:i+T-1}^{t_{0:T-1}}\mid \hat{x}_{i:i+T-1}^{t_{1:T}},\hat{x}_{<i}^{0})
  =
  \Psi\bigl(G_{\theta}(\hat{x}_{i:i+T-1}^{t_{1:T}}, t_{1:T}, \hat{x}_{<i}^{0}), t_{0:T-1}\bigr).
\end{equation}
The front is almost clean after several steps, and the back is still nearly Gaussian as it has just entered.
The window then rolls forward and a new Gaussian chunk is appended.
Causality is relaxed inside the window, so frames stay in order.
Errors can be revised while noisy, so less enters it.

\subsection{Archive \& Working Banks}
\label{sec:method:memory}

\noindent
\begin{minipage}[t]{0.48\textwidth}
Rolling Forcing keeps a global sink and a recent FIFO.
The sink anchors palette and exposure, while the FIFO carries local motion for recent chunks.
After a chunk leaves the FIFO it is in neither, so a person who leaves and later returns, or a scene that cuts away and comes back, is no longer addressable.
The rolling window can still repair neighbors in one pass, but it does not store what happened in between, the mid-horizon gap we change.
We close that gap inside the KV cache the rolling model already writes.
A few-step generator already commits clean keys as it rolls, so the question is not whether to cache but how those keys are kept after they have left the recent window.
We do not add a side store, and we do not cut the first frames into a longer prefix that pretends mid-horizon content is still nearby.
\end{minipage}\hfill
\begin{minipage}[t]{0.50\textwidth}
\vspace{-4em}
\setlength{\abovecaptionskip}{0pt}
\begin{algorithm}[H]
\caption{Archive and Working Step}
\label{alg:memory}
\footnotesize
\begin{algorithmic}[1]
\Require $C=[\mathrm{sink}\mid\mathrm{archive}\mid\mathrm{working}]$
\Require clean chunk $x$ and query window $Q$
\If{sink is not full}
  \State write $\mathrm{KV}(x)$ to sink
\Else
  \If{working is full}
    \State $y \leftarrow \arg\min_{z\in\mathrm{working}}\mathrm{sim}(z,x)$
    \Statex \hfill \Comment{\eqref{eq:promote}}
    \State move $y$ from working to archive
    \If{archive is full}
      \State FIFO-evict the oldest archive chunk
    \EndIf
  \EndIf
  \State append $\mathrm{KV}(x)$ to the working tail
\EndIf
\State $K_{\mathrm{ctx}} \leftarrow [\mathrm{sink};\mathrm{archive};\mathrm{working};\mathrm{current}]$
\State apply RoPE \Comment{\S\ref{sec:method:rope}}
\State \Return $\mathrm{Attend}(Q,K_{\mathrm{ctx}})$
\end{algorithmic}
\end{algorithm}
\end{minipage}

\vspace{-8pt}
As shown in \Figref{fig:memory}, the cache is kept as $C=[\mathrm{sink}\mid\mathrm{archive}\mid\mathrm{working}]$ and split into three banks by role, not by a time cut of the first frames.
Sink still holds the opening chunk and stays fixed, so the origin does not drift.
Working holds recent history so the tail stays continuous.
Archive holds intermediate events that working later had to give up, so mid-horizon stays diverse.

A new clean chunk is written into working, except while the sink is still being filled, and a full sink does not move.
The working tail keeps the newest keys next to the next frames, which local motion needs.
If working is full, dropping an arbitrary recent chunk would break that continuity, and archive should not collect near-duplicates of $x$.
We therefore move
\begin{equation}
\label{eq:promote}
  y=\arg\min_{z\in\mathrm{working}}\mathrm{sim}(z,x)
\end{equation}
from working into archive, where $\mathrm{sim}$ compares working keys to $x$.
That promotion makes archive a diverse mid-horizon store.
If archive is full, FIFO drops the oldest chunk to bound the cache, and the new chunk sits at the working tail.
\Algref{alg:memory} does this write and returns $\mathrm{Attend}(Q,K_{\mathrm{ctx}})$.

\begin{figure}[t]
  \centering
  \includegraphics[width=0.9\linewidth]{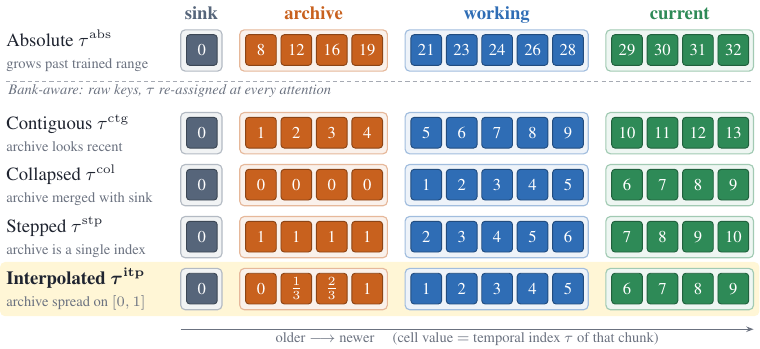}
  \vspace{-4mm}
  \caption{Bank-aware RoPE assigns $\tau$ on one window.
  Absolute $\tau^{\mathrm{abs}}$ lets later indices grow past the trained range.
  Contiguous $\tau^{\mathrm{ctg}}$ numbers from $0$.
  Collapsed $\tau^{\mathrm{col}}$ puts sink and archive at $0$.
  Stepped $\tau^{\mathrm{stp}}$ puts sink at $0$, archive at $1$, working from $2$.
  Interpolated $\tau^{\mathrm{itp}}$ places archive in $(0,1)$.
  }
  \label{fig:tau}
  \vspace{-4mm}
\end{figure}

\subsection{Bank-aware RoPE}
\label{sec:method:rope}

Video transformers encode order with three-dimensional rotary embeddings~\citep{su2024rope}.
We retarget the time axis.
A query at index $\tau$ and a key at index $\tau'$ are rotated by $\tilde{q}=R(\tau)\,q$, $\tilde{k}=R(\tau')\,k$, where $R$ is the rotary block, and $\tilde{q}^{\top}\tilde{k}=q^{\top}R(\tau-\tau')\,k$ depends only on the relative offset, so attention can learn shift-invariant patterns.
In the usual cache rotation is applied at write time, so the stored key is already $\tilde{k}=R(\tau)\,k$.
Later steps reuse that key and do not apply a new $\tau$.
Once written, the phase stays fixed until eviction and the key stays bound to the write-time index.
Values stay unrotated, so moving that key later means storing it raw and rotating it at attention time.

Rolling Forcing makes one exception for the sink: those keys are stored raw and rotated only when replayed just before $L_{\mathrm{tem}}$, and the rest still follows write-time rotation.
Every other key is written with \textbf{Absolute} time $\tau^{\mathrm{abs}}$.
Past the training horizon these $\tau^{\mathrm{abs}}$ indices grow without bound and leave the trained range, so later keys attend under untrained phases and cannot be moved, and only the sink can be placed again.
That binding is worse after three banks: archive keys are older than $L_{\mathrm{tem}}$, so their $\tau^{\mathrm{abs}}$ indices drift farther, and re-encoding only the sink leaves archive on stale indices.
They also need a mid-horizon band that $\tau^{\mathrm{abs}}$ cannot provide, so that RoPE cannot be inherited as is.

Therefore, we store raw keys for the whole window and re-assign $\tau$ at every attention.
Relative time encodings are prior tools~\citep{infinityrope2026}, and attaching them to the three-bank cache is the contribution.
We still have to decide how sink, archive, working, and current sit on the axis.
As shown in \Figref{fig:tau}, we design the four assignments ($N_a$ and $N_w$ count archive and working frames).
\begin{itemize}
  \item \textbf{Contiguous} $\tau^{\mathrm{ctg}}$.
  The aim is to keep every index inside the trained range as one relative clip.
  We concatenate sink, archive, working, and current in time order, number them from the left, and fix the sink at $0$ with no gap, as if the three banks were one FIFO.
  \begin{equation}
  \label{eq:tau-ctg}
    \tau^{\mathrm{ctg}}=(0,1,\ldots,N{-}1).
  \end{equation}
  The indices never leave the trained range and the rule is cheap to apply.
  The three banks then share one scale, so archive sits next to the sink and looks like a moment ago.
  \item \textbf{Collapsed} $\tau^{\mathrm{col}}$.
  The aim is to pin the opening shot and park mid-horizon keys on that origin.
  Every sink and archive token is given index $0$, working starts at $1$ and increases by one frame, and current continues after working.
  Archive has no internal order on the axis.
  \begin{equation}
  \label{eq:tau-col}
  \left\{
  \begin{aligned}
    \tau^{\mathrm{col}}(\mathrm{Sink}) &= 0, \\
    \tau^{\mathrm{col}}(\mathrm{Archive}_{a}) &= 0
    \quad (a=1,\ldots,N_a), \\
    \tau^{\mathrm{col}}(\mathrm{Working}_{w}) &= w
    \quad (w=1,\ldots,N_w).
  \end{aligned}
  \right.
  \end{equation}
  The origin stays fixed, working still has a local order, and all indices remain small.
  A mid-horizon event cannot be told from the opening shot because both sit at $0$.
  \item \textbf{Stepped} $\tau^{\mathrm{stp}}$.
  The aim is to mark three timescales without growing indices.
  The sink sits at $0$, archive at $1$, and working at $2,3,\ldots$, with current after working.
  Neighboring banks differ by one and do not share a slot, but archive is still a single index.
  \begin{equation}
  \label{eq:tau-stp}
  \left\{
  \begin{aligned}
    \tau^{\mathrm{stp}}(\mathrm{Sink}) &= 0, \\
    \tau^{\mathrm{stp}}(\mathrm{Archive}_{a}) &= 1
    \quad (a=1,\ldots,N_a), \\
    \tau^{\mathrm{stp}}(\mathrm{Working}_{w}) &= w{+}1
    \quad (w=1,\ldots,N_w).
  \end{aligned}
  \right.
  \end{equation}
  The banks no longer collide and the indices stay small.
  Archive is still a single point, so frames inside archive cannot be ordered on the axis.
  \item \textbf{Interpolated} $\tau^{\mathrm{itp}}$.
  The aim is to put the sink at $0$, working near the present, and archive in between with its frames distinguishable.
  The $N_a$ archive frames are spread evenly on $[0,1]$, so neighboring archive keys differ by $1/(N_a{-}1)$ and collapse to $0$ if $N_a{=}1$.
  Working increases from $1$ as a near-present chain, and current follows working.
  \begin{equation}
  \label{eq:tau-itp}
  \left\{
  \begin{aligned}
    \tau^{\mathrm{itp}}(\mathrm{Sink}) &= 0, \\
    \tau^{\mathrm{itp}}(\mathrm{Archive}_{a}) &= \frac{a{-}1}{N_a{-}1}
    \quad (a=1,\ldots,N_a,\ 0\text{ if }N_a{=}1), \\
    \tau^{\mathrm{itp}}(\mathrm{Working}_{w}) &= w
    \quad (w=1,\ldots,N_w).
  \end{aligned}
  \right.
  \end{equation}
  The sink stays at the origin, working stays near the present, and archive occupies the mid-horizon band with its frames distinguishable.
\end{itemize}

\begin{figure}[t]
  \centering
  \includegraphics[width=0.8\linewidth]{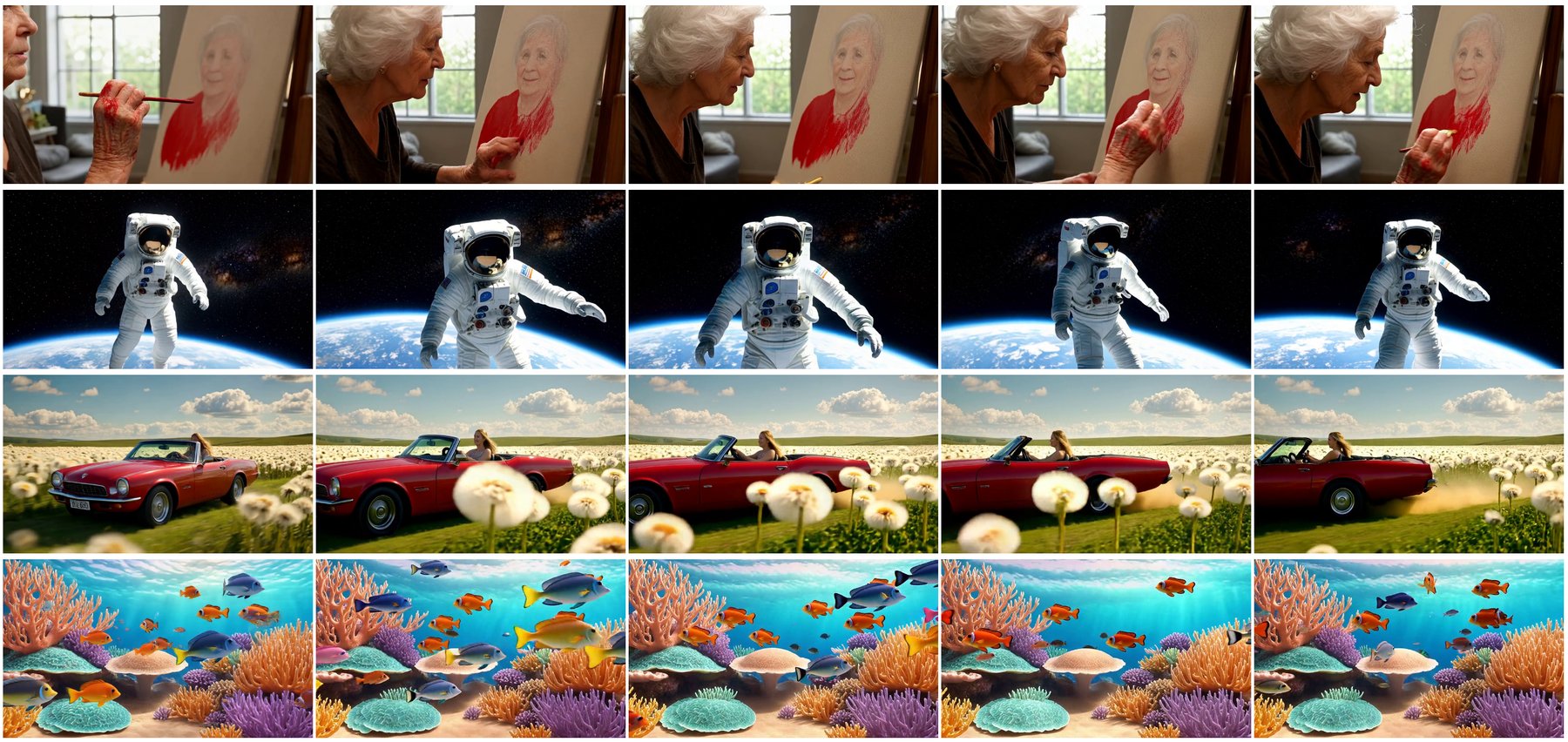}
  \vspace{-3mm}
  \caption{Further qualitative samples from Memory Forcing.
  Each row is one generated clip, frames in time order.
  Fine action and wide scenes keep identity and layout across the clip.}
  \label{fig:qual-ours}
  \vspace{-5mm}
\end{figure}

\subsection{Training}
\label{sec:method:train}

We keep Rolling Forcing's distillation loop, with DMD and a flow critic on non-overlapping windows (\S\ref{sec:method:prelim}, \S\ref{sec:method:rolling}).
A rolling-only rollout assembles the clean video from frames that left different noise levels, so the DMD clip is uneven in clarity.
Training mixes rolling and self-forcing windows with equal probability.
Self-forcing windows regularize camera motion, and inference uses the rolling window alone.
\Algref{alg:memory} now selects RoPE and appends a key, so each step practices the three-bank write and Bank-aware RoPE, and a later clean chunk uses that same write.

If that insertion attends only the tokens already in the cache, the stored keys are formed under a different context than denoising.
We therefore use \textbf{Aligned cache insertion}.
The insertion forward looks at the sink, the extracted archive and working banks, and the current window, the same set that a denoising step attends.
Keys that enter the cache then match the attention used for the loss.

\subsection{Scaling to 5B}
\label{sec:method:scale}

Public few-step autoregressive long-video models stop at Wan2.1-T2V-1.3B~\citep{wan2025}, so we attach the same recipe to Wan2.2 5B~\citep{wan2025} to test whether the three-bank cache and Bank-aware RoPE scale on a stronger backbone.
The banks, Interpolated $\tau^{\mathrm{itp}}$, mixed training, Aligned cache insertion, and DMD stay as in \S\ref{sec:method:memory}--\ref{sec:method:train}.
The 5B cut, cache, and attention budget use that backbone's tokens per frame, and both scales are primary results in \S\ref{sec:exp}.

Wan2.2 5B begins as a bidirectional backbone.
A forcing student is causal, so we first convert that backbone with the first two stages of CausalForcing~\citep{causalforcing2026}, \textbf{autoregressive diffusion training} and \textbf{causal ODE initialization}.
The first stage trains a multi-step causal teacher under teacher forcing, so frame $i$ at noise $t$ sees a clean prefix $x_{<i}^{0}$ as in \S\ref{sec:method:prelim}.
The resulting PF-ODE is injective at the frame level.
The second stage samples trajectories from that teacher and trains the few-step student to regress the clean frame from the noisy state on the same prefix,
\begin{equation}
\label{eq:causal-ode}
\theta^{*}
=
\arg\min_{\theta}
\mathbb{E}
\bigl\|
G_{\theta}(x_i^{t}, x_{<i}^{0}, t) - x_i^{0}
\bigr\|_{2}^{2},
\end{equation}
where $(x_i^{t},x_i^{0})$ lie on the teacher's PF-ODE.
After these two stages the 5B student is causal.
We then train it with the Memory Forcing recipe of \S\ref{sec:method:memory}--\ref{sec:method:train}.

Both stages supervise a next-frame map given clean history, so clip motion sets the target size.
Low-motion clips make that map nearly the identity, so AR conversion is easy but the student copies the last frame.
High-motion clips enlarge the residual and ODE-target variance, so the same regression is hard to fit.
We therefore filter Koala-36M~\citep{koala36m2025} on quality and aesthetics, and mix low, mid, and high motion at $4{:}3{:}3$, yielding $280{,}000$ clips for AR diffusion training and $36{,}000$ for causal ODE initialization, and the 5B result in \S\ref{sec:exp} supports this mix and that the recipe scales.

\section{Experiments}
\label{sec:exp}

We report Memory Forcing at Wan2.1-T2V-1.3B and at Wan2.2 5B as two primary results.
Ablations that isolate the partition, $\tau$, or eviction stay on 1.3B.

\subsection{Implementation Details}
\label{sec:exp:impl}

\paragraph{Model.}
We implement Memory Forcing with Wan2.1-T2V-1.3B~\citep{wan2025} as the default 5s, 16 FPS, $832 \times 480$ base, and with Wan2.2-TI2V-5B~\citep{wan2025} as a second primary result.
We adopt Causal Forcing~\citep{causalforcing2026} weights from two stages, autoregressive diffusion training and causal ODE initialization.
Text prompts are drawn from the filtered and LLM-augmented VidProM~\citep{vidprom2023} dataset.
We set $T{=}4$, 3 latent frames per chunk, and denoising steps $[1000,750,500,250]$.
Training uses $1{,}000$ steps, batch size $32$ on $32$ A800 GPUs, and a window of $33$ latent frames.
The KV cache is partitioned into sink, archive, and working banks of $3$, $6$, and $12$ frames.
We use Bank-aware RoPE with default Interpolated $\tau^{\mathrm{itp}}$, and AdamW for both the generator $G_{\theta}$ (learning rate $1.5 \times 10^{-6}$) and the fake score $s_{\mathrm{gen}}$ (learning rate $4.0 \times 10^{-7}$).
The generator is updated every $5$ steps of fake score updates.

\paragraph{Evaluation.}
We evaluate on four prompt suites: official VBench~\citep{vbench2024} and MovieGen~\citep{moviegen2024}, plus two harder sets, 20 VBench prompts as VBench-Complex and 30 LLM prompts as Gen-Complex.
These fifty prompts emphasize multi-subject scenes, person-to-person interaction, and other mid-horizon difficulties.
We generate each prompt at four durations, 5s (21 latent frames), 15s (63), 30s (120), and 60s (240), all at 16 FPS and $832 \times 480$.
We score videos with the VBench toolkit, reporting Total, Quality, and Semantic on official VBench and Quality only on MovieGen, VBench-Complex, and Gen-Complex.

\subsection{Comparisons}
\label{sec:exp:cmp}

Table~\ref{tab:main} compares Memory Forcing with recent open streaming and long-video models on the four prompt suites and four durations in \S\ref{sec:exp:impl}.
The grid is meant to be read across settings, not as a single 5s contest.
Where a method is publicly released on more than one initialization, we report both the Self-Forcing and CausalForcing weights so that a gap is not an artifact of the starting checkpoint.
Although Memory Forcing is not the strongest row at 5s, it stays at the top of the table on the longer clips, and the drop from 5s to 60s is the smallest among methods that report all four lengths.
That pattern is meant to show that mid-horizon history remains usable after the short window.

\begin{table}[t]
\caption{Comparison on official VBench Total and Quality on VBench-Complex, Gen-Complex, and MovieGen at 5s, 15s, 30s, and 60s.
${CF~ODE}$ and ${SF~ODE}$ are the CausalForcing and Self-Forcing pretraining.
Red is best and blue is second-best in each column.}
\vspace{-3mm}
\label{tab:main}
\begin{center}
\resizebox{\linewidth}{!}{%
\begin{tabular}{lcccccccccccccccccccc}
\toprule
 & \multicolumn{5}{c}{VBench (Total)} & \multicolumn{5}{c}{VBench-Complex (Quality)} & \multicolumn{5}{c}{Gen-Complex (Quality)} & \multicolumn{5}{c}{MovieGen (Quality)} \\
\cmidrule(lr){2-6}\cmidrule(lr){7-11}\cmidrule(lr){12-16}\cmidrule(lr){17-21}
Method & 5s & 15s & 30s & 60s & $\Delta^{\text{5s}\rightarrow\text{60s}}_{\downarrow}$ & 5s & 15s & 30s & 60s & $\Delta^{\text{5s}\rightarrow\text{60s}}_{\downarrow}$ & 5s & 15s & 30s & 60s & $\Delta^{\text{5s}\rightarrow\text{60s}}_{\downarrow}$ & 5s & 15s & 30s & 60s & $\Delta^{\text{5s}\rightarrow\text{60s}}_{\downarrow}$ \\
\midrule
CausVid & 82.99 & \na{} & \na{} & \na{} & \na{} & 80.83 & \na{} & \na{} & \na{} & \na{} & 80.35 & \na{} & \na{} & \na{} & \na{} & 79.62 & \na{} & \na{} & \na{} & \na{} \\
NOVA & 78.54 & 74.78 & \na{} & \na{} & \na{} & 73.32 & 69.18 & \na{} & \na{} & \na{} & 74.86 & 69.41 & \na{} & \na{} & \na{} & 74.08 & 68.86 & \na{} & \na{} & \na{} \\
SkyReels-V2 & 82.65 & 80.11 & \na{} & \na{} & \na{} & 81.08 & 75.07 & \na{} & \na{} & \na{} & 78.83 & 71.93 & \na{} & \na{} & \na{} & 79.88 & 75.35 & \na{} & \na{} & \na{} \\
Self-Forcing & 84.09 & \na{} & \na{} & \na{} & \na{} & 81.37 & \na{} & \na{} & \na{} & \na{} & 79.99 & \na{} & \na{} & \na{} & \na{} & 81.09 & \na{} & \na{} & \na{} & \na{} \\
LongLive & 83.23 & 82.95 & 82.10 & 82.21 & -1.02 & 79.73 & 80.01 & 79.07 & 78.81 & -0.92 & 81.51 & 80.59 & 79.98 & \textcolor{red}{80.36} & \textcolor{red}{-1.15} & 80.15 & 79.79 & 79.20 & 79.13 & \textcolor{red}{-1.02} \\
LongLive~$_{{inf}}$ & 82.97 & 82.84 & 82.43 & 81.77 & -1.20 & 79.45 & 80.70 & 80.45 & 80.37 & \textcolor{red}{0.92} & \textcolor{blue}{81.57} & \textcolor{blue}{80.81} & \textcolor{red}{80.52} & 79.61 & -1.96 & 80.14 & 79.93 & 79.52 & 78.82 & \textcolor{blue}{-1.32} \\
SANA-Video & 83.18 & 83.17 & 82.80 & 82.22 & \textcolor{blue}{-0.96} & 80.18 & 78.63 & 78.44 & 77.29 & -2.89 & \textcolor{red}{83.27} & 80.37 & 80.46 & 78.59 & -4.68 & 80.91 & 80.31 & 79.22 & 78.00 & -2.91 \\
RollingForcing~$_{{SF~ODE}}$ & 83.63 & 82.82 & 82.04 & 81.64 & -1.99 & 80.23 & 80.83 & 79.00 & 78.59 & -1.64 & 81.48 & 78.83 & 77.77 & 78.36 & -3.12 & 80.21 & 79.07 & 78.61 & 77.85 & -2.36 \\
RollingForcing~$_{{CF~ODE}}$ & 83.43 & 80.33 & 78.61 & 77.24 & -6.19 & \textcolor{blue}{82.38} & 81.16 & 80.51 & 79.12 & -3.26 & 80.18 & 77.95 & 77.88 & 76.75 & -3.43 & \textcolor{red}{82.87} & 80.91 & 79.64 & 78.68 & -4.19 \\
InfinityStar & 84.03 & 75.94 & 66.76 & 61.47 & -22.56 & 80.34 & 71.45 & 62.31 & 58.86 & -21.48 & 79.71 & 72.59 & 63.36 & 59.10 & -20.61 & 80.03 & 70.75 & 62.19 & 60.19 & -19.84 \\
Infinity-RoPE~$_{{SF~ODE}}$ & 83.78 & 82.78 & 81.58 & 79.99 & -3.79 & 81.04 & 80.59 & 79.74 & 79.18 & -1.86 & 81.15 & 79.44 & 78.55 & 77.67 & -3.48 & 81.09 & 79.70 & 78.65 & 77.07 & -4.02 \\
Infinity-RoPE~$_{{CF~ODE}}$ & 84.08 & 82.29 & 80.71 & 79.51 & -4.57 & 81.65 & 79.29 & 77.63 & 77.40 & -4.25 & 80.13 & 77.95 & 77.22 & 76.88 & -3.25 & \textcolor{blue}{82.42} & 80.30 & 79.21 & 78.11 & -4.31 \\
RewardForcing & \textcolor{red}{84.21} & \textcolor{red}{83.63} & \textcolor{blue}{83.03} & 82.06 & -2.15 & 82.06 & \textcolor{blue}{81.55} & \textcolor{blue}{81.20} & \textcolor{blue}{80.73} & -1.33 & 81.26 & 79.77 & 79.47 & 79.36 & \textcolor{blue}{-1.90} & 81.56 & 80.96 & \textcolor{blue}{80.63} & \textcolor{blue}{80.19} & -1.37 \\
DeepForcing~$_{{SF~ODE}}$ & 84.08 & 83.18 & 82.37 & \textcolor{blue}{82.37} & -1.71 & 81.05 & 80.54 & 80.29 & 79.89 & -1.16 & 80.00 & 78.81 & 78.50 & 77.88 & -2.12 & 81.05 & 79.93 & 78.57 & 78.17 & -2.88 \\
DeepForcing~$_{{CF~ODE}}$ & 84.05 & 82.42 & 80.55 & 80.24 & -3.81 & 81.75 & 79.90 & 77.28 & 77.22 & -4.53 & 80.39 & 78.02 & 75.70 & 75.44 & -4.95 & 82.36 & 80.05 & 77.87 & 76.71 & -5.65 \\
CausalForcing & 83.90 & \na{} & \na{} & \na{} & \na{} & 81.84 & \na{} & \na{} & \na{} & \na{} & 80.56 & \na{} & \na{} & \na{} & \na{} & 82.36 & \na{} & \na{} & \na{} & \na{} \\
AnchorForcing & \textcolor{blue}{84.16} & 83.26 & 82.76 & 81.73 & -2.43 & 80.92 & 81.19 & 80.74 & 80.41 & -0.51 & 80.39 & 78.53 & 78.45 & 78.15 & -2.24 & 81.30 & \textcolor{blue}{80.97} & 80.47 & 79.81 & -1.49 \\
\midrule
Memory Forcing & 83.88 & \textcolor{blue}{83.51} & \textcolor{red}{83.58} & \textcolor{red}{83.20} & \textcolor{red}{-0.68} & \textcolor{red}{82.41} & \textcolor{red}{82.13} & \textcolor{red}{81.89} & \textcolor{red}{81.99} & \textcolor{blue}{-0.42} & 81.07 & \textcolor{red}{81.54} & \textcolor{blue}{80.48} & \textcolor{blue}{79.92} & \textcolor{red}{-1.15} & 82.28 & \textcolor{red}{81.55} & \textcolor{red}{80.98} & \textcolor{red}{80.66} & -1.62 \\
Memory Forcing (5B) & 84.24 & 83.83 & 83.11 & 81.93 & -2.31 & 81.34 & 80.69 & 80.43 & 79.77 & -1.57 & 80.16 & 79.87 & 80.15 & 79.19 & -0.97 & 81.34 & 80.63 & 79.91 & 79.16 & -2.18 \\
\bottomrule
\end{tabular}
}
\end{center}
\vspace{-5mm}
\end{table}

\begin{table}[t]
\caption{Ablations on Wan2.1-T2V-1.3B~\citep{wan2025} scored on official VBench~\citep{vbench2024} at 30s and 60s.
Every panel reports Total/Quality/Semantic at 16 FPS and $832 \times 480$.}
\vspace{-3mm}
\label{tab:ablate}
\centering
\begin{subtable}[t]{0.32\linewidth}
\centering
\caption{Memory banks.}
\label{tab:ablate-mem}
\resizebox{\linewidth}{!}{\begin{tabular}{lccc}
\toprule
Memory & Total & Quality & Semantic \\
\midrule
Official & 82.04 & 82.64 & 79.61 \\
Causal-Forcing checkpoint & 78.61 & 81.30 & 67.85 \\
Working & 82.41 & 83.46 & 78.22 \\
\rowcolor{gray!15}
Archive + working & 83.58 & 84.57 & 79.63 \\
\bottomrule
\end{tabular}
}
\end{subtable}
\hfill
\begin{subtable}[t]{0.32\linewidth}
\centering
\caption{Bank-aware RoPE.}
\label{tab:ablate-rope}
\resizebox{\linewidth}{!}{\begin{tabular}{lccc}
\toprule
RoPE & Total & Quality & Semantic \\
\midrule
Contiguous $\tau^{\mathrm{ctg}}$ & 81.31 & 82.37 & 77.10 \\
Stepped $\tau^{\mathrm{stp}}$ & 82.56 & 83.37 & 79.29 \\
Collapsed $\tau^{\mathrm{col}}$ & 81.50 & 82.41 & 77.87 \\
\rowcolor{gray!15}
Interpolated $\tau^{\mathrm{itp}}$ & 83.58 & 84.57 & 79.63 \\
\bottomrule
\end{tabular}
}
\end{subtable}
\hfill
\begin{subtable}[t]{0.32\linewidth}
\centering
\caption{Similarity eviction.
}
\label{tab:ablate-evict}
\resizebox{\linewidth}{!}{\begin{tabular}{lccc}
\toprule
Write & Total & Quality & Semantic \\
\midrule
FIFO (30s) & 83.39 & 84.59 & 78.59 \\
\rowcolor{gray!15}
Similarity (30s) & 83.58 & 84.57 & 79.63 \\
\midrule
FIFO (60s) & 82.96 & 83.94 & 79.05 \\
\rowcolor{gray!15}
Similarity (60s) & 83.20 & 84.11 & 79.57 \\
\bottomrule
\end{tabular}
}
\end{subtable}
\vspace{-8mm}
\end{table}

\subsection{Ablations}
\label{sec:exp:ablate}


\begin{figure}[t]
  \centering
  \includegraphics[width=0.9\linewidth]{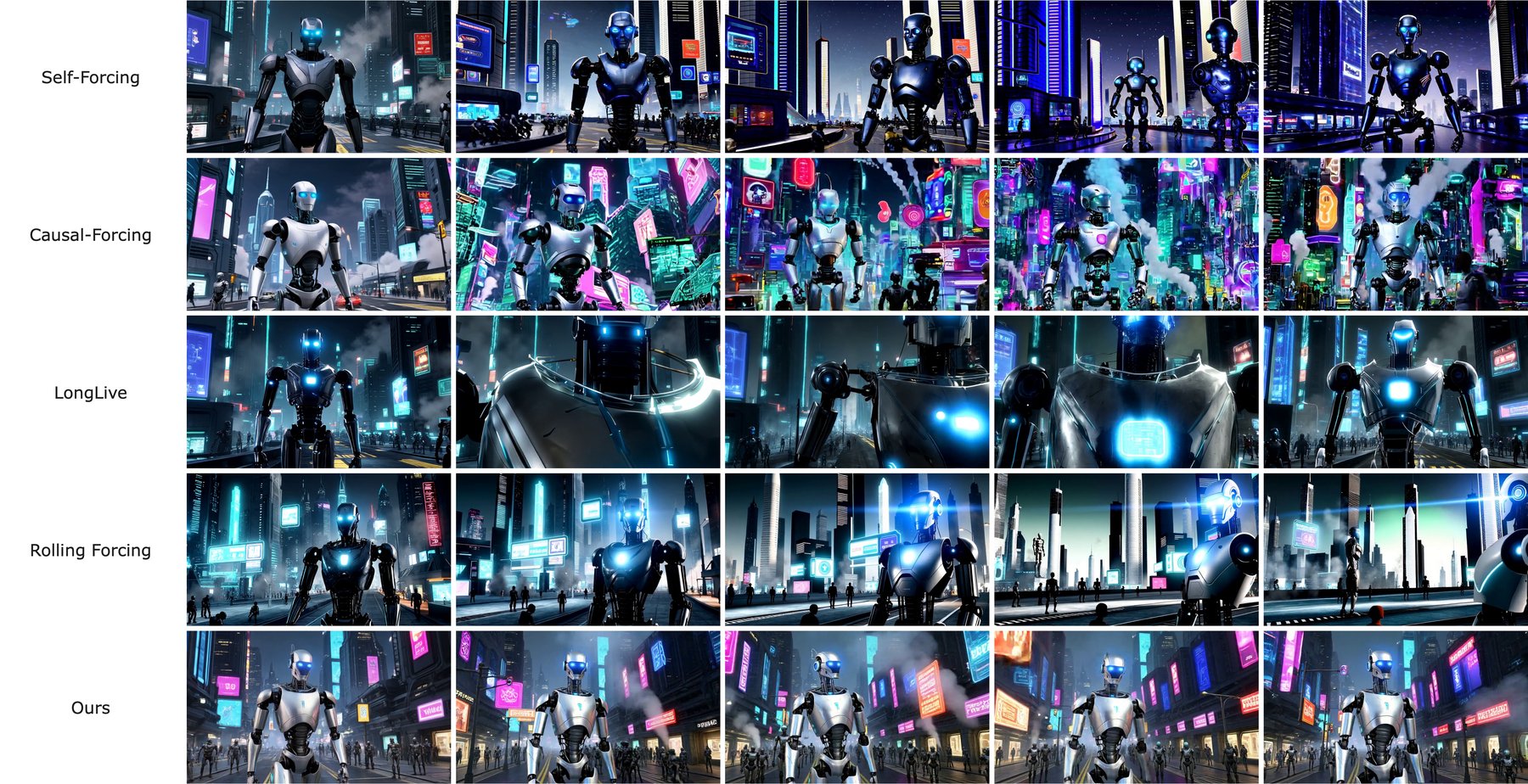}
  \vspace{-2.5mm}
  \caption{Qualitative comparison at 1.3B against four recent open-source models.
  Memory Forcing keeps the subject's identity and the scene layout, while the baselines drift or collapse.}
  \label{fig:qual13}
  \vspace{-4mm}
\end{figure}

\begin{figure}[t]
  \centering
  \includegraphics[width=0.9\linewidth]{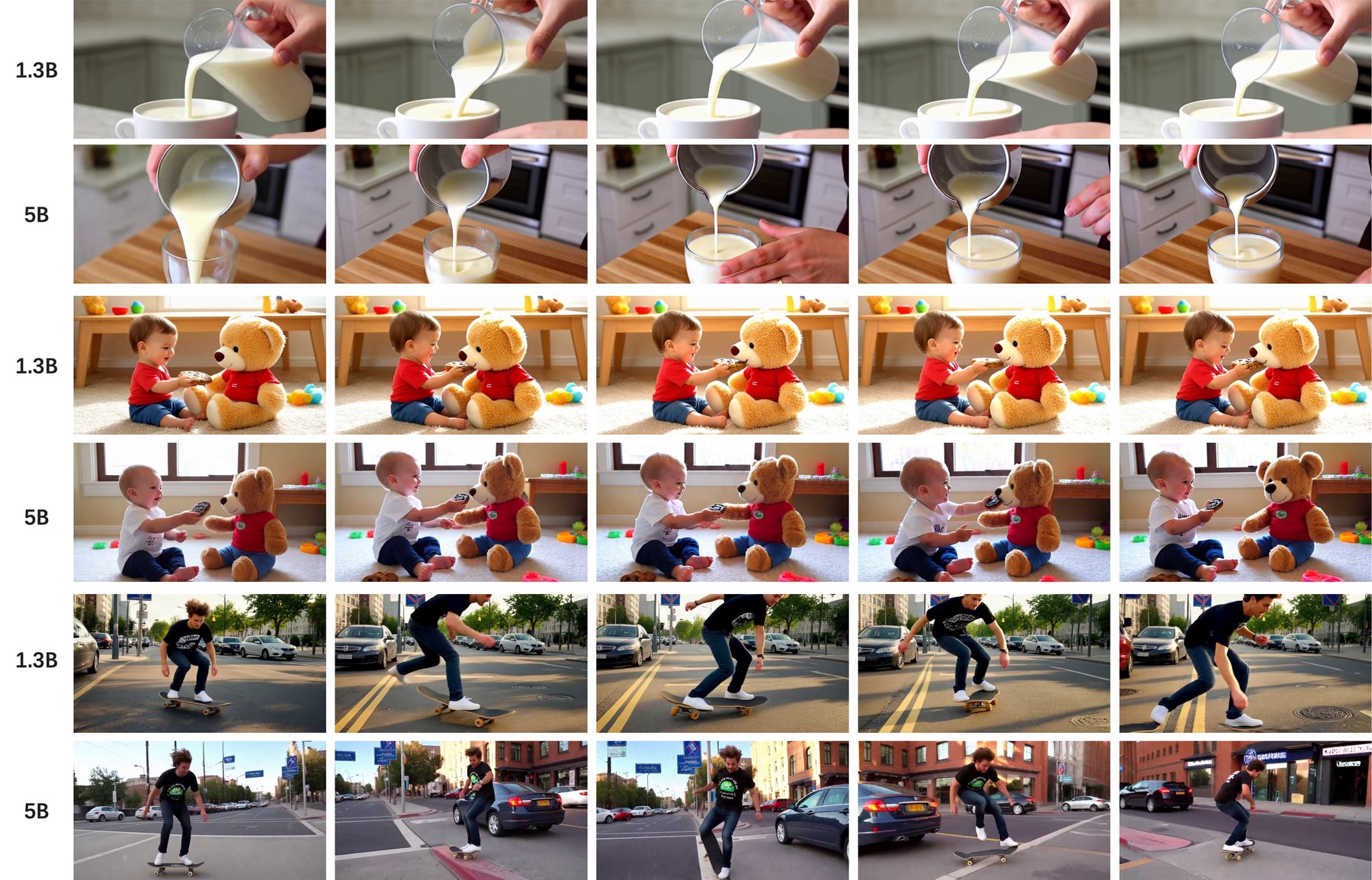}
  \vspace{-2.5mm}
  \caption{Memory Forcing at Wan2.1-T2V-1.3B and Wan2.2 5B on the same prompts.
  The 5B run is more physically plausible, more realistic, and more dynamic.}
  \vspace{-6mm}
  \label{fig:qual5b}
\end{figure}

\paragraph{Memory banks.}
Table~\ref{tab:ablate-mem} changes only the cache: official Rolling Forcing, a Causal-Forcing checkpoint, a working bank alone, and archive plus working.
Archive plus working is the strongest row.
Working alone is lower on Total, Quality, and Semantic, so the extra bank is a gain.
Official Rolling Forcing sits below it, and the Causal-Forcing checkpoint is lower still, especially on Semantic.
A dedicated mid-horizon bank therefore keeps events that a flat window drops.

\paragraph{Bank-aware RoPE.}
Table~\ref{tab:ablate-rope} freezes the three-bank cache and swaps only the index map: Contiguous $\tau^{\mathrm{ctg}}$, Stepped $\tau^{\mathrm{stp}}$, Collapsed $\tau^{\mathrm{col}}$, and default Interpolated $\tau^{\mathrm{itp}}$.
Interpolated $\tau^{\mathrm{itp}}$ is the strongest row.
Stepped is second.
Contiguous and Collapsed fall further, especially on Semantic.
Archive history has to sit on $(0,1)$ with frames that remain distinguishable; packing it with the present, collapsing it onto the sink, or stepping the whole bank onto one integer all weaken retrieval.

\paragraph{Similarity eviction.}
Table~\ref{tab:ablate-evict} freezes the three-bank cache and $\tau^{\mathrm{itp}}$ and changes only which working chunk is written into archive.
Similarity is the training default.
FIFO is used only at inference, where it moves the oldest working chunk into archive.
Similarity is the stronger of the two rows on Total, Quality, and Semantic.
FIFO turns archive into a second recency queue, while Similarity moves the working chunk least like the incoming event so mid-horizon slots stay diverse.

\subsection{Qualitative results}
\label{sec:exp:qual}

We use qualitative frames to show Memory Forcing on its own samples, then in a 1.3B comparison, and then at Wan2.2 5B on harder scenes.
\Figref{fig:qual-ours} shows Memory Forcing clips on diverse prompts, with identity and layout held across each clip.
\Figref{fig:qual13} compares Memory Forcing at 1.3B with four recent open-source models.
Frames are aligned across time on the same prompt.
The baselines drift or collapse the subject's identity and the scene layout.
Memory Forcing keeps both, which is the point of keeping an archive.
\Figref{fig:qual5b} then shows the same method at Wan2.2 5B.
On crowded and multi-subject scenes the 5B run keeps objects and interactions that the 1.3B rollout starts to lose.

\section{Conclusion}
\label{sec:conc}

We present Memory Forcing, a few-step streaming method for retrieving events after they leave the recent attention window.
Its Archive \& Working Banks retain a diverse mid-horizon history within a fixed cache, while Bank-aware RoPE gives the three banks distinct temporal positions and keeps that history accessible as generation proceeds.
Together, these changes make long rollouts markedly more stable without growing attention cost with clip length.
At 1.3B, Memory Forcing leads on longer clips, shows the smallest drop from 5s to 60s among methods with complete results, and preserves subjects and scenes through leave-and-return.
We further show that the same design scales to Wan2.2 5B, producing more physically plausible, realistic, and dynamic videos and, to our knowledge, the first public 5B model on this forcing line.

\bibliography{iclr2027_conference}
\bibliographystyle{iclr2027_conference}


\end{document}